\documentclass[wcp]{jmlr}

\usepackage{longtable}% for long tables

\usepackage{booktabs}
\title[Deep Competitive Pathway Networks]{Deep Competitive Pathway Networks}

  \author{\Name{Jia-Ren Chang} \Email{followwar.cs00g@nctu.edu.tw}\\
  \Name{Yong-Sheng Chen} \Email{yschen@cs.nctu.edu.tw}\\
  \addr Department of Computer Science, National Chiao Tung University, Hsinchu, Taiwan}

\begin{document}

\maketitle

\begin{abstract}
In the design of deep neural architectures, recent studies have demonstrated the benefits of grouping subnetworks into a larger network. 
For examples, the Inception architecture integrates multi-scale subnetworks and the residual network can be regarded that a residual unit combines a residual subnetwork with an identity shortcut.
In this work, we embrace this observation and propose the Competitive Pathway Network (CoPaNet).
The CoPaNet comprises a stack of competitive pathway units and each unit contains multiple parallel residual-type subnetworks followed by a max operation for feature competition.
This mechanism enhances the model capability by learning a variety of features in subnetworks.
The proposed strategy explicitly shows that the features propagate through pathways in various routing patterns, which is referred to as pathway encoding of category information.
Moreover, the cross-block shortcut can be added to the CoPaNet to encourage feature reuse.
We evaluated the proposed CoPaNet on four object recognition benchmarks: CIFAR-10, CIFAR-100, SVHN, and ImageNet.
CoPaNet obtained the state-of-the-art or comparable results using similar amounts of parameters. The code of CoPaNet is available at: \url{https://github.com/JiaRenChang/CoPaNet}.
\end{abstract}

\section{Introduction}
Deep convolutional neural networks (CNNs) have been shown to be highly effective in image classification with large datasets, such as CIFAR-10/100~\citep{krizhevsky2009learning}, SVHN~\citep{netzer2011reading}, and ImageNet~\citep{deng2009imagenet}.
Improvements in computer hardware and network architectures have made it possible to train deeper and more complex networks.
%AlexNet~\cite{krizhevsky2012imagenet} had 5 convolutional layers, VGG~\cite{simonyan2014very} had 19 layers, and Inception~\cite{szegedy2015rethinking} had 22 layers.

Network grouping is an efficient technique to improve the accuracy in model learning.
The Inception architecture~\citep{szegedy2015going} was proposed to aggregate abundant features via multi-scale subnetworks.
In addition, dueling architecture~\citep{wang2015dueling} in deep reinforcement learning can explicitly exploit subnetworks to represent state value and action advantages.
Recently, the Residual Networks (ResNets)~\citep{he2015deep,he2016identity} can be regarded that a residual unit includes an identity shortcut and a residual subnetwork.
%ResNets~\cite{he2015deep,he2016identity} used identity shortcut connections, for which their outputs were added to the output of the stacked layers, thereby forcing the stacked layers to approximate the residual functions.
This approach can alleviate the vanishing gradient problem by bypassing the gradients without attenuation and thus can increase the network depth up to more than 100 layers.
As suggested in~\citep{abdi2016multi,huang2016deep,veit2016residual}, ResNets gains its superior performance by implicitly averaging many subnetworks.
%Further improvements of network architecture are on increasing the capacity of ResNets.

The redundancy problem of ResNets has been raised in~\citep{huang2016deep, zagoruyko2016wide}.
Some studies primarily aimed at the improvement of the propagation in ResNet, thereby reducing the redundancy problem.
Stochastic Depth~\citep{huang2016deep} tackled this problem by randomly disabling residual units during training.
%This technique prevented the co-adaptation of resdiual units and thus improved the capacity of ResNet.
Wide Residual Networks~\citep{zagoruyko2016wide} addressed this problem by decreasing the depth and increasing the width of residual units for faster training. 
%These networks have advantages that the wider residual units can offer higher capacity and shallow architecture keeps propagation continuing.
Both of these network architectures are attempts to shorten the network and thereby improve information back-propagation during training. 
Without shortening network, a recent work~\citep{he2016identity} analyzed various usages of rectified linear unit (ReLU) and batch normalization (BN) in ResNets for direct propagation, and proposed methods for identity mapping in residual units to improve training in very deep ResNets. 

Some studies encouraged the direct feature reuse by replacing the element-wise addition in ResNets with concatenation.
FractalNet~\citep{larsson2016fractalnet} repeatedly combines many subnetworks in a fractal expansion rule to obtain large nominal network depth.
DenseNet~\citep{huang2016densely} is similar to FractalNet with the difference that DenseNet connects each layer to all of its preceding layers.
These approaches exhibit a behavior of mimicking deep supervision, which is important to the learning of discriminative features.

Some studies aimed at the improvement of the residual units by representing the residual function with many tiny subnetworks.
Inception-ResNet~\citep{szegedy2016inception} presented Inception-type residual units.
PolyNet~\citep{zhang2016polynet} replaces the original residual units with polynomial combination of Inception units for enhancing the structural diversity. 
Multi-residual networks~\citep{abdi2016multi} and ResNeXt~\citep{xie2016aggregated} both aggregate residual transformations from many tiny subnetworks.

\begin{figure}[tbp]
\begin{center}
	\includegraphics*[width=\linewidth]{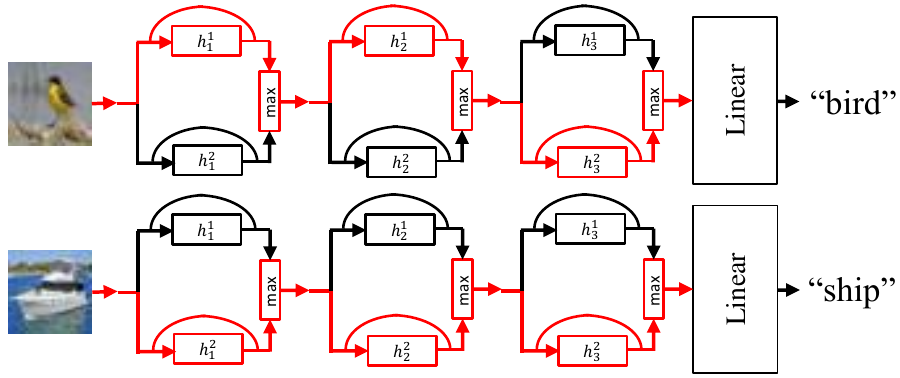}
\end{center}
  \caption{The concept of \textit{pathway encoding} in the proposed architecture. The category information is encoded on the route (red arrows) through which features propagate.}
\label{copanet}
\end{figure}

The idea behind the use of subnetworks is to simplify network for efficient training.
By explicitly factoring the network into a series of operations, features can be learned independently.
In this work, we embrace this observation and propose a novel deep architecture referred to as Competitive Pathway Network (CoPaNet).
Because the underlying mapping function can be decomposed into the maximum of multiple simpler functions and the residual learning~\citep{he2015deep} is a good strategy for approximating the mapping functions, the proposed competitive pathway (CoPa) unit was designed to comprise multiple parallel residual-type subnetworks followed by a max operation for feature competition.
Furthermore, identity cross-block shortcuts can be added to the CoPaNet to enhance feature reuse.
These strategies offer several advantages: 
1. Feature redundancy can be reduced by dropping unimportant features through competition.
2. The competitive mechanism facilitates the network to modularize itself into multiple parameter-sharing subnetworks for parameter efficiency~\citep{srivastava2013compete}.
3. CoPaNet uses residual-type subnetworks and therefore inherits the advantage of ResNet for training very deep network.
4. With competitive mechanism and residual-type subnetworks, the CoPaNet explicitly exhibits the property of pathway encoding, as shown in Figure \ref{copanet}. Because the residual-type subnetwork can preserve feature identity such that the winning path can be traced back within the entire network.
That is, the \textit{routing pattern of propagating features} encodes category information.
5. The cross-block shortcuts encourage coarse feature reuse and implicit deep supervision.

%In contrast to separating residual function into many tiny subnetworks as in~\cite{szegedy2016inception,xie2016aggregated}, we first factorized the model into several subnetworks (we called \textit{pathways}) and then added a shortcut to each of those subnetworks to force them to approximate the residual functions.
%We suppose that those subnetworks are 
%CoPaNet is established by stacking multiple competitive pathway units. 
%Each unit may include several residual-type pathways in parallel, which compete with each other.
%With competition and residual-type pathway, CoPaNet explicitly exhibits the behavior of \textit{pathway encoding}, as shown in Figure \ref{copanet}.
%Because the residual-type pathways can preserve feature identity such that the winning pathways can be traced back within the entire network, the \textit{routing pattern of propagating features} encodes category information.

CoPaNet was evaluated using several benchmark datasets such as CIFAR-10, CIFAR-100, SVHN, and ImageNet. 
Our resulting models performed equally to or better than the state-of-the-art methods on the above-mentioned benchmark datasets.

\section{Related Work}

\subsection{Residual Networks (ResNets)}

ResNets~\citep{he2015deep} are motivated by the counterintuitive observation that the performance of neural networks actually gets worse when developed to a very great depth. 
This problem can be attributed to the fact that the gradient vanishes when information back-propagates through many layers. 
~\cite{he2015deep} proposed skipping some of the layers in convolutional networks through the implementation of shortcut connections, in the formulation of an architecture referred to as residual units. The original residual unit performs the following computation:
\begin{align*} x_{l+1}=ReLU(id(x_l )+f_l (x_l )) \, , \end{align*}
where $x_l$ denotes the input feature of the $l$-th residual unit, $id(x_l)$ performs identity mapping, and $f_l$ represents layers of the convolutional transformation of the $l$-th residual unit.

~\cite{he2016identity} further suggested to replace ReLU with another identity mapping, allowing the information to be propagated directly. 
Thus, they proposed a pre-activation residual unit with the following form:
\begin{align*}x_{l+1}=id(x_l )+f_l (x_l ) \, . \end{align*}
Furthermore, the positions of BN and ReLU are changed to allow the gradients to be back-propagated without any transformation.
Their experimental results demonstrated the high efficiency of pre-activation residual units.

\subsection{Competitive Nonlinear functions}
Maxout Networks~\citep{goodfellow2013maxout} were recently introduced to facilitate optimization and model averaging via Dropout. The authors of this work proposed a competitive nonlinearity referred to as maxout, which was constructed by obtaining the maximum across several maxout hidden pieces. Maxout Networks can be regarded as universal approximators and can provide better gradient back-propagation than other activation functions.
Without down-sampling the features, Local Winner-Take-All (LWTA)~\citep{srivastava2013compete} was inspired by the characteristics of biological neural circuits. 
Each LWTA block contains several hidden neurons and produces an output vector determined by local competition between hidden neurons activations.
Only the winning neuron retains its activation, whereas other hidden neurons are forced to shut off their activation.
In empirical experiments, both network architectures have been shown to have advantages over ReLU.

\begin{figure*}
\begin{center}
	\includegraphics*[width=6in]{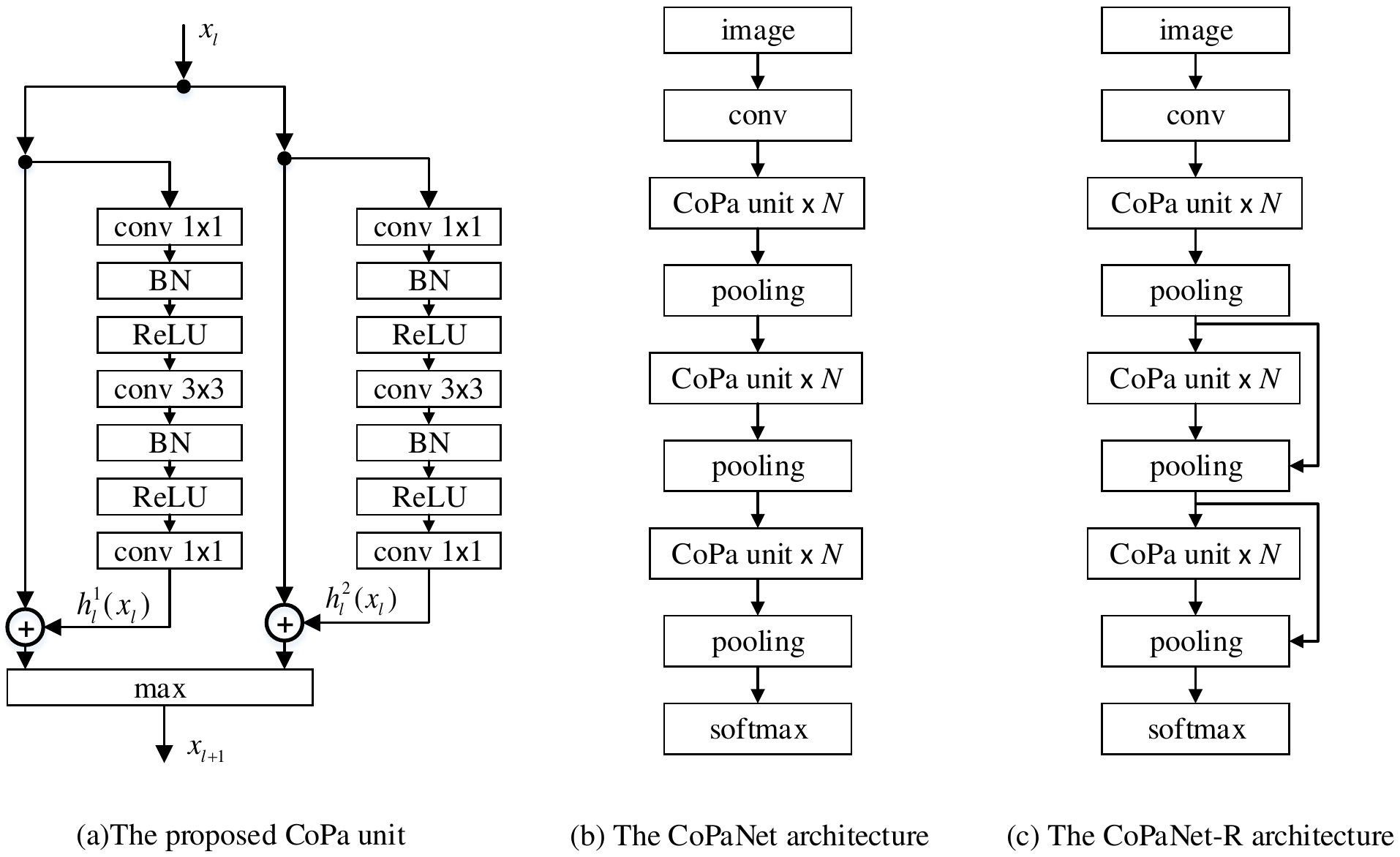}
\end{center}
   \caption{The proposed CoPa unit and network architecture.}
\label{resunit}
\end{figure*}

\section{Competitive Pathway Network}
\subsection{Competitive pathway unit}
CoPaNet is an attempt to separate model into subnetworks through competition.
In the following, we refer to residual-type subnetworks as pathways.
In a CoPa unit, multiple pathways are compiled in parallel and features are selected by using a $\max$ operation.
A CoPa unit includes output $x_{l+1}$ with $K$ pathways, which can be formulated as follows:
\begin{align*}
x_{l+1}=\mathop{\max }_{k\in [1,\ldots , K]} z_l^k,
\end{align*}
where $z_l^k=id(x_l)+h_l^k (x_l)$, $x_l$ is the input feature, and $h_l^k (x_l)$ represents layers of transformations on the $k$-th pathway at the $l$-th CoPa unit. Figure \ref{resunit}a illustrates the CoPa unit (featuring two pathways) used in this paper. 

Competitive pathways appear complex; however, the proposed CoPaNet is easy to train. 
Notice that residual learning~\citep{he2015deep} is based on the hypothesis that underlying mapping function $H(x)$ is very hard to fit.
Nevertheless, the mapping function can be decomposed into two simpler functions: $H(x)=x+F(x)$. 
~\cite{he2015deep} claimed that the residual function $F(x)$ is easier to approximate. 
Motivated by the idea of streamlining the process of approximating the underlying mapping function, we first decompose the underlying mapping function into the maximum of two simpler functions, that is, $H(x)=\max\{f(x),g(x)\}$. 
We then use residual learning~\citep{he2015deep} and let $f(x)=x+h^1 (x)$ and $g(x)=x+h^2 (x)$.\
The desired mapping function becomes $H(x)=\max\{x+h^1 (x),x+h^2 (x)\}$. 
This illustrates the need for two parallel networks (one each for $h^1(x)$ and $h^2(x)$), each of which comprises several stacked layers in order to approximate discrete residual functions. 
Because $f(x)$ and $g(x)$ are simpler, it would be easier to approximate $h^1 (x)$ and $h^2 (x)$ than the original residual learning~\citep{he2015deep}.
Our CoPa unit is different from maxout unit~\citep{goodfellow2013maxout}.
The original maxout unit is constructed to obtain the maximum across several elementary neurons.
 Our method replaces the elementary neurons with generic functions, which are modeled by ResNets.

Further, the property of pathway encoding reveals in this architecture.
We consider a 2-pathway (denote as $h^1_l, h^2_l$) CoPaNet with three stacked CoPa units, as show in Figure~\ref{copanet}.
We denote that the output of the first CoPa unit is $y_1 = x + h^1_1(x)$ (if $h^1_1$ wins) where $x$ is the input feature.
The output of the second CoPa unit can be written as $y_2 = y_1 + h^1_2(y_1)$ (if $h^1_2$ wins).
The output of the third CoPa unit can be written as $y_3 = y_2 + h^2_3(y_2)$ (if $h^2_3$ wins).
The final output actually can be expressed as $y_3 = x + h^1_1(x) + h^1_2(y_1) + h^2_3(y_2)$.
This indicates that the final output is contributed by three winning subnetworks $h^1_1,  h^1_2, h^2_3$ with reference to $x$.
Thus, the routing pattern can be revealed by propagating $x$ through the entire network.

Within a biological context, competitive mechanisms play an important role in attention~\citep{lee1999attention}. 
Researchers formulated a biological computational model in which attention activates a winner-take-all competition among neurons tuned to different visual patterns. 
In this model, attention alters the thresholds used to detect orientations and spatial frequencies. 
This suggested that winner-take-all competition can be used to explain many of the basic perceptual consequences of attention~\citep{lee1999attention}.

\subsection{CoPaNet Architecture}
CoPaNets can be simply constructed by stacking CoPa units. 
Let the opponent factor $k$ denote the number of pathway in a CoPa unit and the widening factor $m$ multiplies the number of features in convolutional layers.
That is, the baseline CoPa unit corresponds to $k = 2$, $m=1$; whereas ResNet corresponds to $k = 1$, $m=1$.

Figure \ref{resunit}b shows the architecture for CIFAR and SVHN as well as Table~\ref{params} detailed the deployment.
The residual shortcut in the proposed network performs identity mapping and the projection shortcut is used only to match dimensions (using 1$\times$1 convolutions) as ResNet~\citep{he2015deep,he2016identity}. 
For each pathway, we adopted a ``bottleneck'' residual-type unit comprising three convolutional layers (1$\times$1, 3$\times$3, 1$\times$1). 
Alternatively, we could select a ``basic'' residual-type unit comprising two convolutional layers (3$\times$3, 3$\times$3). 
In practice, a ``bottleneck'' residual-type unit is deeper than a ``basic'' one, providing higher dimensional features. 
In the proposed CoPaNet, we placed BN and ReLU after all but the last convolutional layer in every pathway. 

\subsection{Cross-block Shortcut}
The cross-block shortcuts were motivated by DenseNet~\citep{huang2016densely} which reused features from all previous layers with matching feature map sizes.
In contrast to DenseNet~\citep{huang2016densely}, we propose a novel feature reuse strategy: to reuse the features from previous CoPa block (stacked by many CoPa units).
This is accomplished by adding identity shortcuts after pooling layers and concatenate with the output of the next block.
We refer to our model with the cross-block shortcuts as CoPaNet-R, as shown in Figure \ref{resunit}c.

\begin{table*}
\begin{center}
\caption{Network architectures for CIFAR/SVHN (left) and ImageNet (right). Parameters of competitive pathway units are presented in braces (see also Figures~\ref{resunit}b and c). Construction parameters for internal pathways are shown in brackets. The number of pathway is determined by the factor $k$ and the network width is determined by the factor $m$.  The numbers in CoPaNet-26/50/101/164 denote the depths of neural network. For the sake of clarity, the final classification layer has been omitted. }
\includegraphics*[width=6in]{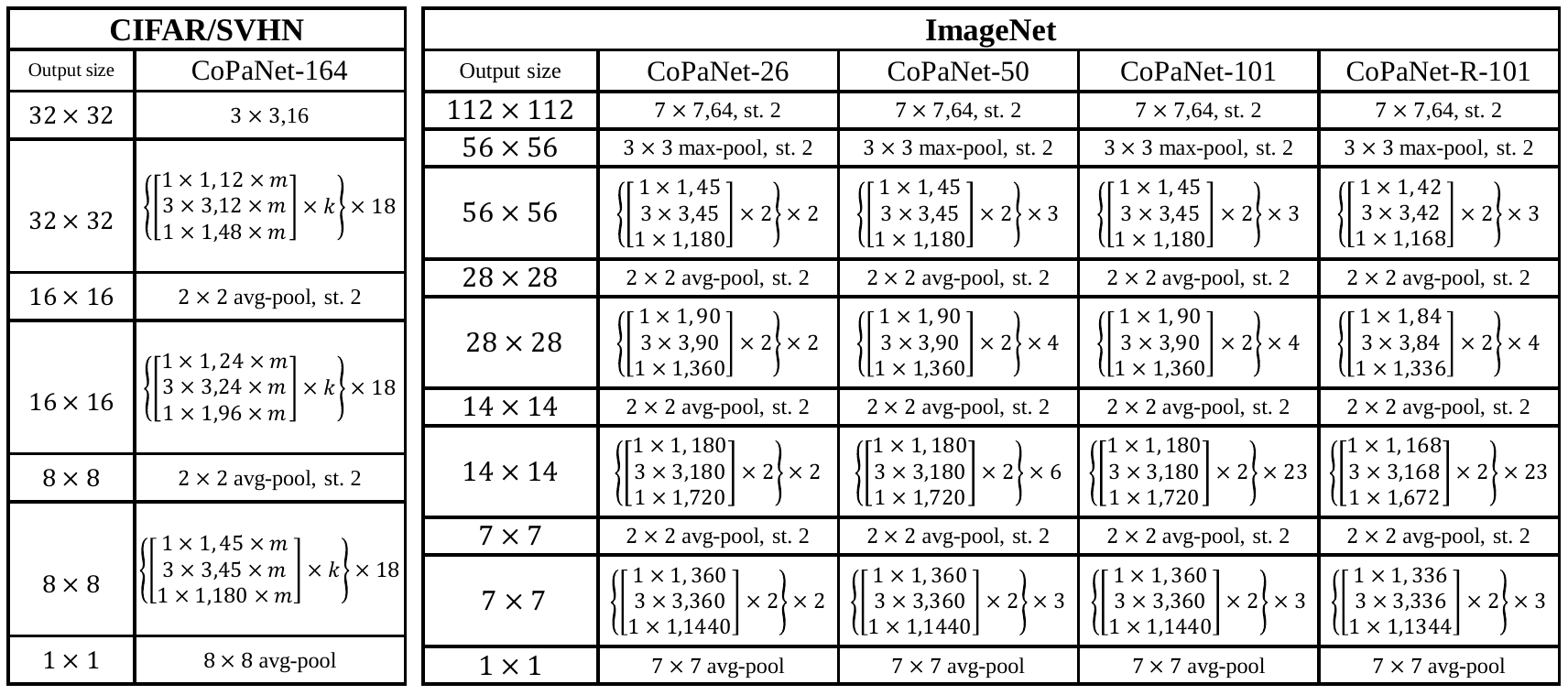}
\label{params}
\end{center}
\end{table*}

\begin{table*}
\centering
   \caption{Comparison of test error on CIFAR and SVHN. The value of $k$ denotes the number of hidden pieces or pathways used in a given competition. The symbol ``+'' indicates data augmentation (translation and horizontal flipping ).}

\begin{center}
\resizebox{1\linewidth}{!}{
\begin{tabular}{lcccccc}
\hline
 Method& Dropout&Depth&Params&C10+&C100+& SVHN  \\ \hline
 Maxout Network ($k$=2)~\citep{goodfellow2013maxout}&${\surd}$& - & - & 9.38 & 38.57 & 2.47  \\
 Network In Network~\citep{DBLP:journals/corr/LinCY13}&${\surd}$& - &0.98 M& 8.81  & 35.68 & 2.35\\
 Maxout Network In Network ($k$=5)~\cite{chang2015batch}&${\surd}$& - &1.6 M& 6.75 & 28.86 & 1.81\\
 Highway Network~\citep{srivastava2015training}&  & - & - & 7.60 & 32.34 & -\\ \hline
 ResNet~\citep{he2015deep}&  & 110 & 1.7 M & 6.43 & - & -\\\hline
Stochastic Depth~\cite{huang2016deep} &  & 110 & 1.7 M & 5.23 & 24.58 & 1.75\\
 &  & 1202 & 19.4 M & 4.91 & - & -\\ \hline
pre-activation ResNet~\citep{he2016identity} &  & 164 & 1.7 M & 5.46 & 24.33 & -\\
 &  & 1001 & 10.2 M & 4.62 & 22.71 & -\\ \hline
Wide ResNet (width=8)~\citep{zagoruyko2016wide} &  & 16 & 11.0 M & 4.27 & 20.43 & -\\
{ }{ }{ }{ }{ }{ }{ }{ }{ }{ }{ }{ }{ }{ }{ }{ }{ }{ }(width=10)&${\surd}$& 28 & 36.5 M & 3.89 & 18.85 & -\\\hline
DenseNet{ }{ }{ }{ }{ } (growth rate=24)~\citep{huang2016densely}&  & 100 & 27.2 M & 3.74 & 19.25 & 1.59 \\
DenseNet-BC (growth rate=40)&  & 190 & 25.6 M & 3.46 & \textbf{17.18} & - \\ \hline
CoPaNet ($k$=2, width=1) &${\surd}$& 164 & 1.75 M & 4.50 & 22.86 & 1.86\\
CoPaNet ($k$=2, width=2)& ${\surd}$ & 164 & 6.98 M & 4.10 & 20.48 & 1.83\\
CoPaNet ($k$=2, width=4) & ${\surd}$ & 164 & 27.9 M & 3.74 & 18.67 & 1.73\\ \hline
CoPaNet-R ($k$=2, width=2) &${\surd}$& 164 & 7.00 M & 3.55 & 20.29 & 1.72\\
CoPaNet-R ($k$=2, width=3)& ${\surd}$ & 164 & 15.7 M & \textbf{3.38} & 18.90 & \textbf{1.58}\\ \hline
\end{tabular}
}
\end{center}

\label{comparison}
\end{table*}

\section{Experiments}
We have tested the proposed CoPaNets and CoPaNets-R on several datasets, and compared the results with those of the state-of-the-art network architectures, especially ResNets.

\subsection{Training}

We constructed a CoPaNet-164, with a set number of pathways ($k=2$), and network width ($m=1,2,4$), detailed in Table~\ref{params}. 
Furthermore, we constructed a CoPaNet-R-164, with a set number of pathways ($k=2$), and network width ($m=2,3$). 

The networks were trained from scratch by using Stochastic Gradient Descent with 300 and 20 epochs for CIFAR and SVHN datasets, respectively. 
The learning rate for CIFAR began at 0.1, divided by 10 at 0.6 and 0.8 fractions of the total number of training epochs. 
The learning rate for SVHN began at 0.1, divided by 10 at 0.5 and 0.75 fractions of the total number of training epochs. 
A batch size of 128 was used for all tests, except for $m=4$ when we used a batch size of 64. 

On ImageNet, we trained from scratch for 100 epochs.
As shown in Table \ref{params}, we constructed several CoPaNets with 2 pathways for ImageNet. 
The learning rate began at 0.1 and was divided by 10 after every 30 epochs.
The model was implemented using Torch7 from the Github repository \textit{fb.resnet.torch} (\url{https://github.com/facebook/fb.resnet.torch}).
Other settings were set exactly the same as those used for ResNet.

We adopted a weight decay of 0.0001 and momentum of 0.9 as in~\citep{he2015deep}.
Weights were initialized in accordance with the methods outlined by ~\cite{he2015delving}. 
We also applied Dropout~\citep{srivastava2014dropout} after the average poolings except the last pooling, and it was deterministically multiplied by (1 - Dropout-rate) at test time.
The Dropout rate was set to 0.2 for CIFAR and SVHN as well as 0.1 for ImageNet. 
The test error was evaluated using the model obtained from the final epoch at the end of training.

\begin{figure*}
\begin{center}
	\includegraphics*[width=6in]{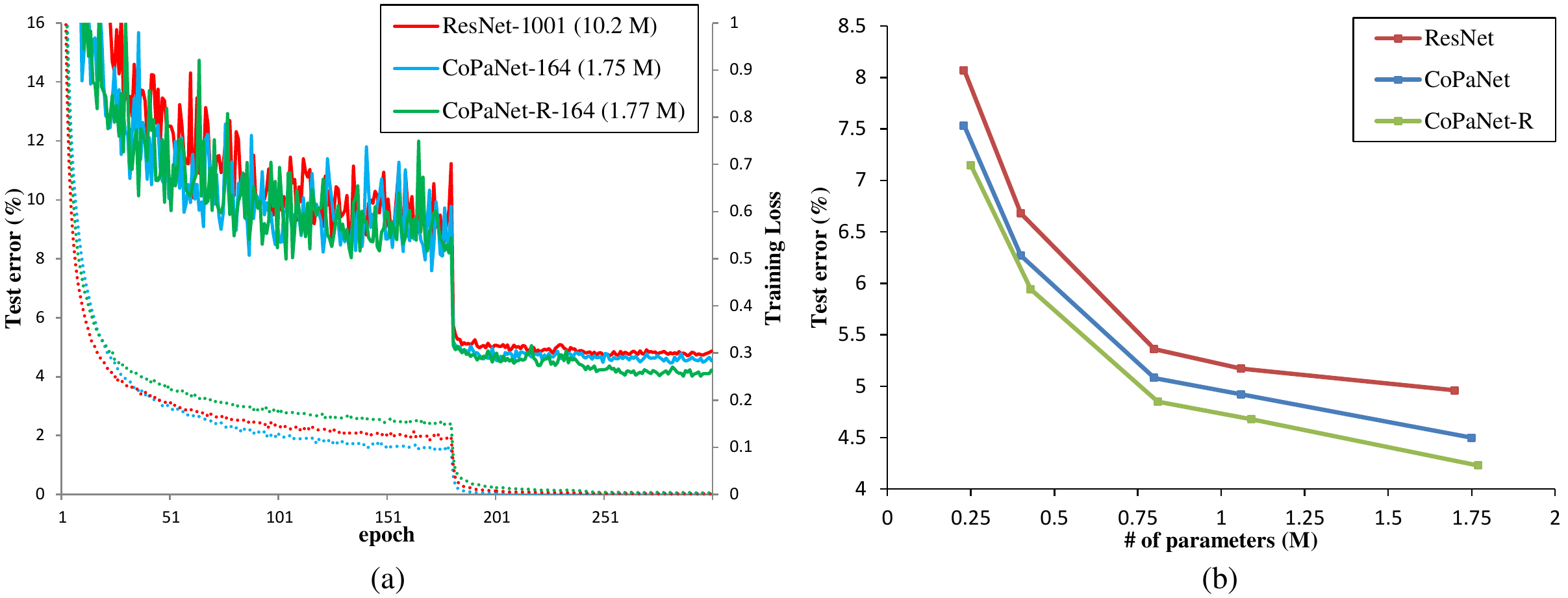}
\end{center}
   \caption{ (a) Training loss (dashed line) and test error (solid line) curves of the pre-activation ResNet-1001 (10.2M), CoPaNet-164 (1.75M), and  CoPaNet-R-164 (1.75M). (b) Comparison of the parameter efficiency between pre-activation ResNets, CoPaNet, and CoPaNet-R.}
\label{curve}
\end{figure*}

\subsection{CIFAR-10}
The CIFAR-10 dataset consists of natural color images, 32$\times$32 pixels in size, from 10 classes, and with 50,000 training and 10,000 test images. Color normalization was performed as data preprocessing. To enable a comparison with previous works, the dataset was augmented by translation as well as random flipping on the fly throughout training. 

As shown in Table \ref{comparison}, we obtained test error of 4.50\%, 4.10\%, and 3.74\% when using network width of $m=1$, $2$, and $4$, respectively. 
We then compared CoPaNet-164 (1.75 M, $m=1$) to pre-activation ResNet-1001 (10.2 M), for which ~\cite{he2016identity} reported test error of 4.62\% (we obtained 4.87\% in our training procedure). 
Figure \ref{curve}a presents a comparison of training and testing curves. 
Furthermore, Our best result on CIFAR-10 was obtained by CoPaNet-R.
We obtained 3.38\% test error with only 15.7 M parameters.

\subsection{CIFAR-100}
The CIFAR-100 dataset is the same size and format as CIFAR-10; however, it contains 100 classes. 
Thus, the number of images in each class is only one tenth that of CIFAR-10. Color normalization was performed as data preprocessing. 
We also performed data augmentation (translation and horizontal flipping) on the CIFAR-100 dataset.

As shown in Table \ref{comparison}, we obtained the test error of 22.86\%, 20.48\%, and 18.67\% for network width of $m=1$, $2$, and $4$ with Dropout, respectively.
CoPaNet-164 (1.75 M, $m=1$) was compared to pre-activation ResNet-164 (1.7 M) for which ~\cite{he2016identity} reported test error of 24.33\%. 
This puts the proposed network on par with pre-activation ResNet-1001 (10.2 M) which achieved test error of 22.71\%.
However, CoPaNet-R showed few benefits on CIFAR-100, and it obtained same level of accuracy.

\subsection{SVHN}
The SVHN dataset consists of color images of house numbers (32$\times$32 pixels) collected from Google Street View. 
This includes 73,257 digits in the training set, 26,032 digits in the test set, and 531,131 in an extra set. 
We used the entire training set and extra set for training. 
We did not perform any data augmentation or preprocessing except for dividing the image intensity by 255.

As shown in Table \ref{comparison}, the CoPaNet-164 (1.75 M, width $m=1$) with test error of 1.86\%. 
CoPaNet-R-164 (width $m=3$) achieved the state-of-the-art results (1.58\%) with only 15.7 M parameters.

\subsection{ImageNet}
\begin{figure}
\centering
\includegraphics*[width=3.6in]{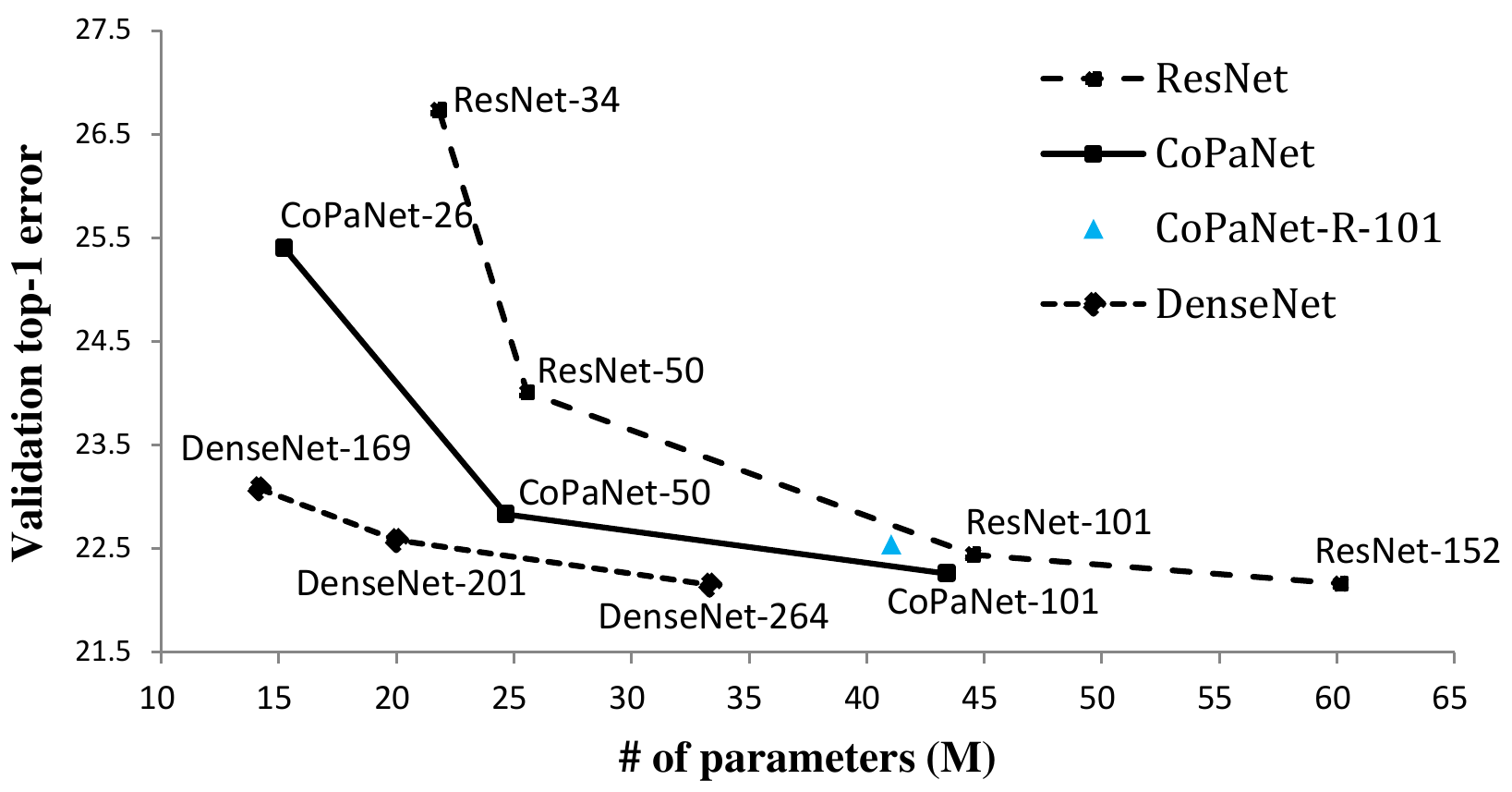}
\caption{The comparison of top-1 validation error (single model and single crop with size 224$\times$224)  across various number of parameters among ResNet, DenseNet, and CoPaNet. }
\label{imagenet}
\end{figure}

The ImageNet 2012 dataset consists of 1000 classes of images with 1.28 millions for training, 50,000 for validation, and 100,000 for testing. 
As shown in Table \ref{params}, we constructed two-pathway CoPaNet with various depths for ImageNet. 
However, we reduce the number of feature maps to approximately 70\% in order to retain a similar number of parameters. 
For a fair comparison, all results were achieved when the crop size was 224$\times$224. 
Our results of single crop top-1 validation error showed better performance than ResNet, as shown in Figure~\ref{imagenet}.
These results reveal that CoPaNets perform on par with the state-of-the-art ResNets, while requiring fewer parameters.
CoPaNets performed worse than DenseNet with similar amounts of parameters.
The major reason could be that DenseNets were much deeper than CoPaNets.

%%%%%%%%%%%%%%%%%%%%%%%%%%%%%%%%
%\begin{figure}[htbp]
%\begin{center}
%	\includegraphics*[width=3in]{Fig_numpath.pdf}
%\end{center}
 %  \caption{The influence of the number of pathways on performance in experiments based on CIFAR-10. More %pathways tends to lower test errors at the expense of more parameters.}
%\label{pathway}
%\end{figure}
%%%%%%%%%%%%%%%%%%%%%%%%%%%%%%%
%\begin{figure}[htbp]
%\begin{center}
%	\includegraphics*[width=3in]{Fig_droprate.pdf}
%\end{center}
%   \caption{Investigation of various Dropout rates on CIFAR dataset.}
%\label{droprate}
%\end{figure}
%%

%%
\section{Discussion}

\subsection{Parameter Efficiency}
The competitive mechanism modularizes the network into multiple parameter-sharing subnetworks and thus can improve parameter efficiency~\citep{srivastava2013compete}.
We trained multiple small networks with various depths on CIFAR-10+.
As shown in Figure~\ref{curve}b, both CoPaNet and its variant outperformed pre-activation ResNet.
The CoPaNet-R achieved better performance than CoPaNet.
When achieving the same level of accuracy, furthermore, CoPaNet requires around a half of the parameters of pre-activation ResNet.

\subsection{Number of Pathways}
\begin{figure}
\centering
\includegraphics*[width=3.6in]{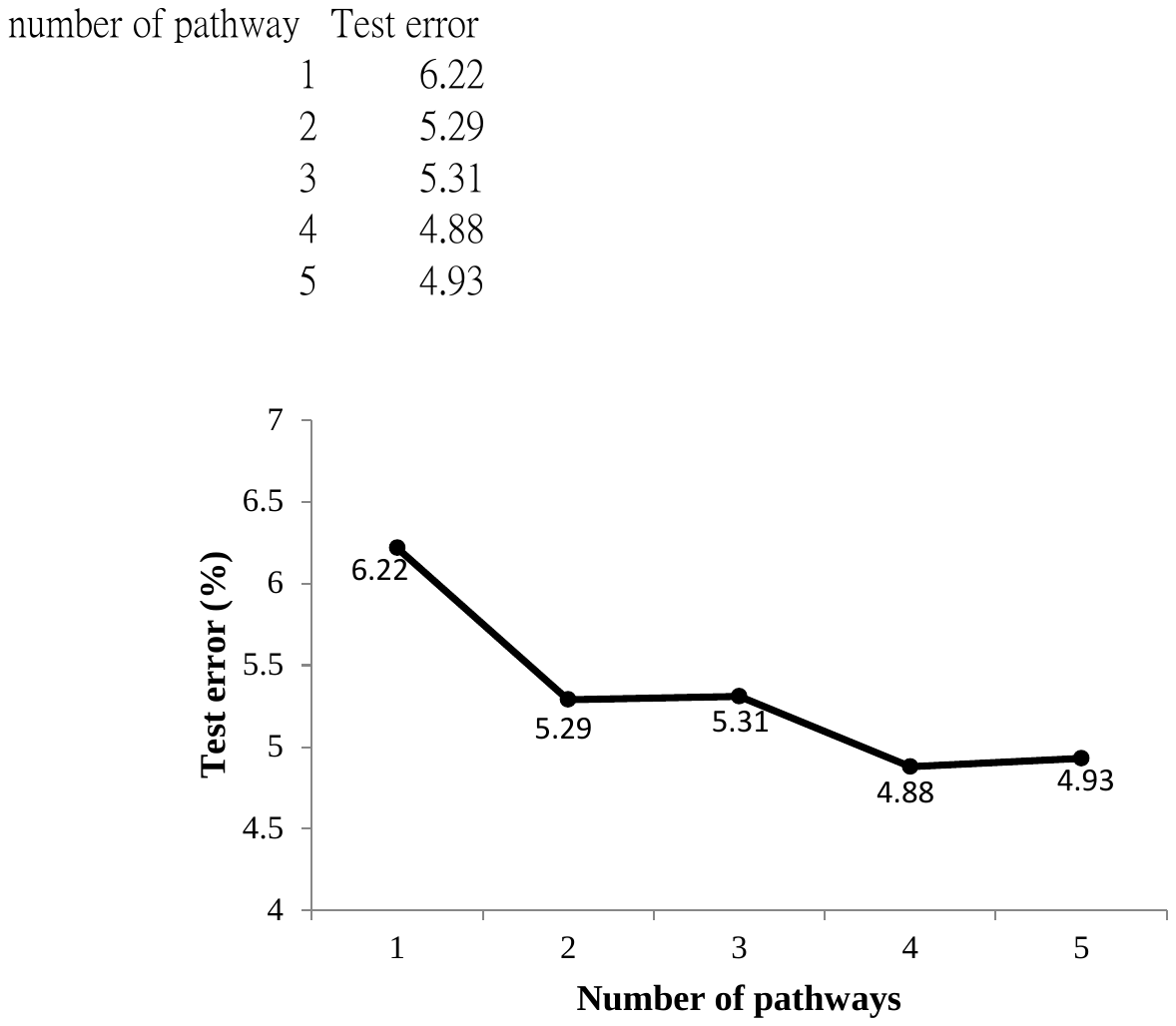}
\caption{The influence of the number of pathways on performance in experiments based on CIFAR-10+. More pathways tends to lower test errors at the expense of more parameters.}
\label{pathway}
\end{figure}

Figure~\ref{pathway} demonstrates that CoPaNet has the capacity to exploit many pathways. 
We trained several CoPaNets-56 (width $m=1$) for use on CIFAR-10+ using various numbers of pathways with the Dropout rate set to 0.2. 
As shown in Figure \ref{pathway}, CoPaNet gains its benefit by increasing the number of pathways to handle complex dataset.
More pathways tend to lower test errors at the expense of more parameters.
Nonetheless, we adopted two pathways in our experiments to restrict the number of parameters.

\subsection{Pathway Encoding}
One paper~\citep{srivastava2014understanding} argued that ReLU network can also encode on subnetwork activation pattern, such as maxout and LWTA networks.
~\cite{srivastava2014understanding} discussed about the activation pattern of many filters in the same layer.
In contrast to~\cite{srivastava2014understanding}, we demonstrated the routing pattern that \textit{one feature map propagate through many stacked pathways (subnetworks)}.

\begin{figure*}
\begin{center}
	\includegraphics*[width=6in]{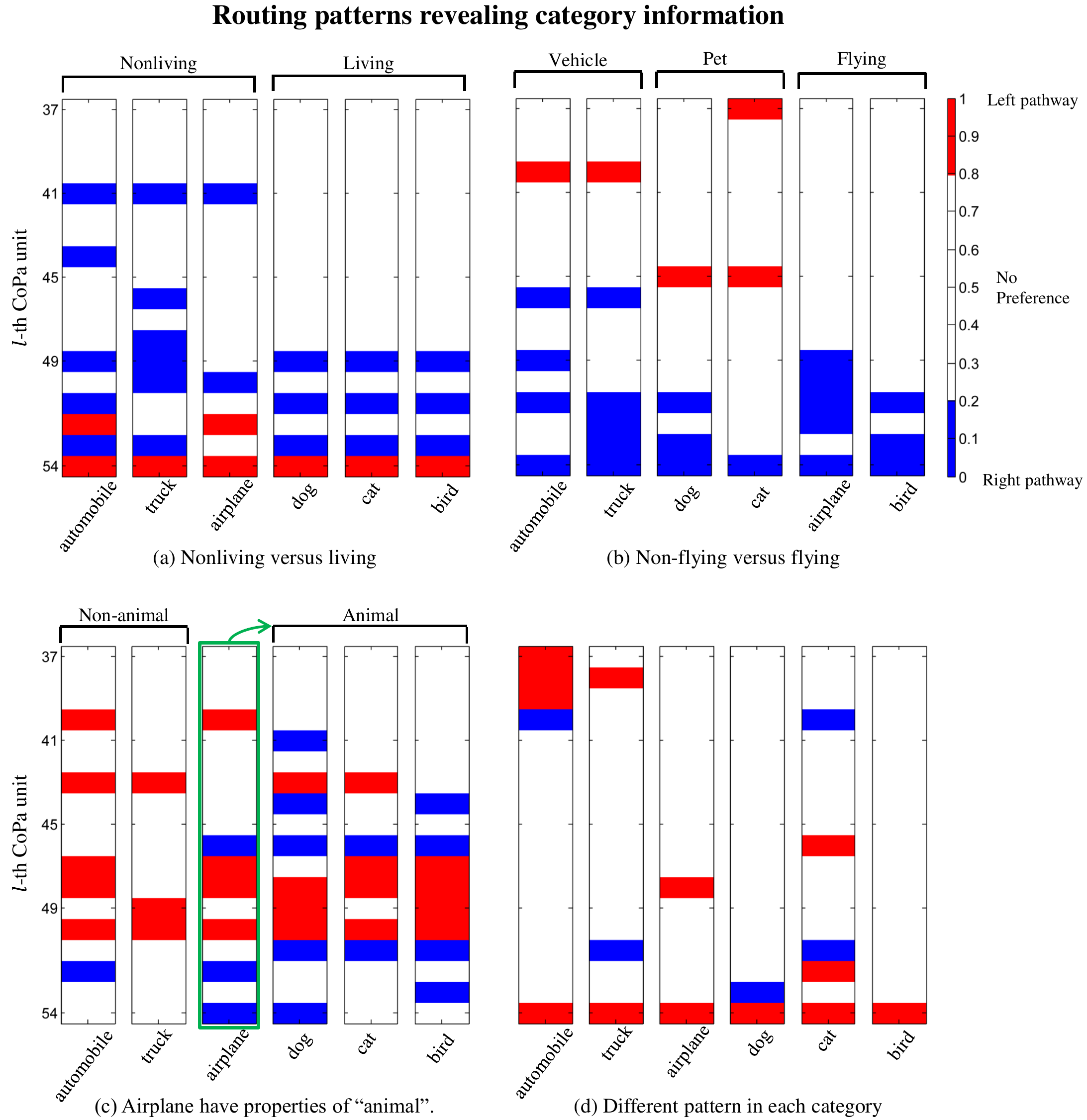}
\end{center}
   \caption{Routing patterns showing the preference of pathway selection in a trained 2-pathway CoPaNet-164 for the CIFAR-10 test dataset. Red color denotes a preference for the left pathway, blue color for the right pathway, and white color for no preference. The vertical axis denotes the $l$-th CoPa units, where $l$ indicates the depth.  The category information can be represented by the routing pattern, which is referred to as  \textit{pathway encoding} in the proposed work. Each sub-figure denotes the routing pattern that one feature map propagates through its preferred route in the network.
Routing patterns between (a) non-living vs. living, (b) non-flying vs. flying, (c) non-animal vs. animal, and (d) different categories are illustrated. Notice that the airplane category shows the routing pattern of ``bird'' in the ``animal'' group.}
\label{pathwayencode}
\end{figure*}

We suppose that the routing patterns are similar within the same semantics and are different between distinct semantics, which is termed as pathway encoding.
As shown in Figure~\ref{pathwayencode}, we calculated the preference of routing patterns in a trained 2-pathway CoPaNet-164 (width $m=1$). 
The preference of pathway was statistically estimated from the CIFAR-10 test set and can reveal the characteristics of the category.
We illustrates the routing patterns in the last block (comprising 18 CoPa units) which contained high-level features.
Each sub-figure showns the routing pattern of one feature map (4 representative feature maps were manually selected from the total of 180), and the color denoted the preference of pathways.
As shown in Figure~\ref{pathwayencode}a, a selected routing pattern can be regarded as encoding the non-living or living groups and the routing patterns are similar in the same group. 
Figure~\ref{pathwayencode}b illustrates that the routing pattern may be encoding the flying concept such that the routing patterns of airplanes are similar to those of birds.
Notice that although airplanes belong to non-living group, there exists a special pattern resembling those of animals, including the bird, as shown in Figure~\ref{pathwayencode}c. 
Furthermore, Figure~\ref{pathwayencode}d illustrates the diversity of routing patterns for different categories.
The similarity and diversity support our hypothesis that CoPaNet is able to use pathway encoding to well represent the object images of different groups. 

\subsection{Coarse Feature Reuse}
\begin{figure}
\centering
\includegraphics*[width=4in]{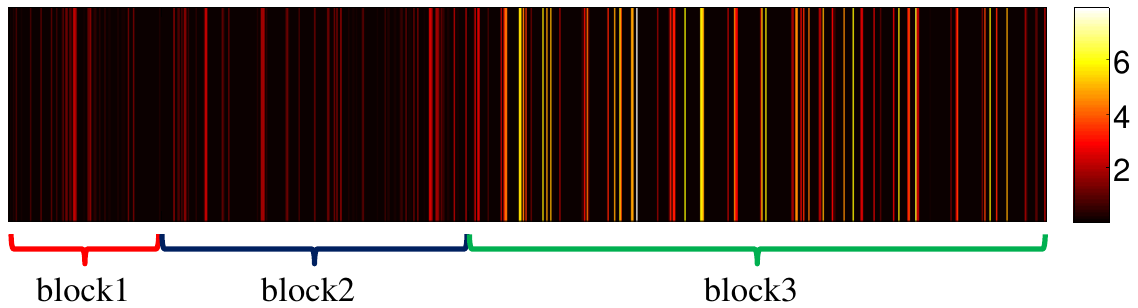}
\caption{The color-encoded $L^1$-norm of the weights of the last classification layer. Notice that the last classification layer  concatenates outputs from all of the three CoPa blocks through cross-block shortcuts.}
\label{reuse}
\end{figure}

The CoPaNet-R architecture adds identity cross-block shortcuts to encourage feature reuse.
This facilitates that the last classification layer can reuse coarse features from all previous blocks.
Thus those shortcuts provide additional supervision because classifiers are attached to every CoPa blocks.
We trained a CoPaNet-R-164 (width $m=2$) on CIFAR-10+ and it achieved 3.55\% test error, as shown in Table~\ref{comparison}.
Figure~\ref{reuse} shows the $L^1$-norm of weights of the last classification layer. 
In this figure, we can observe that the last classification layer uses features from early blocks.
The concentration towards the final block suggests that high-level features dominate in classification.

However, CoPaNet-R did not outperform CoPaNet on CIFAR-100 and ImageNet.
This may be due to the relatively few training samples for each class (500 samples per class in CIFAR-100 as well as around 1000 samples per class in ImageNet).
We conducted an experiment to demonstrate this effect.
We used a small CIFAR-10 dataset (1000 training samples per class) to train CoPaNet-164 and CoPaNet-R-164, both with width $m=2$, and achieved test errors of 12.58\% and 12.53\%, respectively.
There is no significant difference in this case.
With full training set (5000 training samples per class), CoPaNet-R has significant improvement compared to CoPaNet, as shown in Table~\ref{comparison}.
The coarse feature reuse may be effective only when the amount of training samples is large enough for each class.

\section{Conclusions}
This paper proposes a novel convolutional neural network architecture, the CoPaNet. 
It introduces a nice property that input features transmit through various routing patterns for different category information, called pathway encoding.
Empirical results demonstrate that the category information plays a role in selecting pathways.
We showed that CoPaNet inherits the advantages of ResNet which can scale up to hundreds of layers.
In our experiments, CoPaNet yielded improvements in accuracy as the number of parameters increased. 
Moreover, CoPaNet requires fewer parameters to achieve the same level of accuracy as the state-of-the-art ResNet.
We  further proposed a novel feature reuse strategy, CoPaNet-R: adding cross-block shortcuts in order to encourage the reuse of output from all previous blocks. 
According to our experiments, CoPaNet-R can learn accurate models by exploiting the reuse of coarse features.

Our study showed that network partitioning, feature competition, and feature reuse can lead to performance improvements.
CoPaNet and its variant obtained the state-of-the-art or competitive results on several image recognition datasets.
Other studies showed that competitive networks have other beneficial properties such as mitigation of catastrophic forgetting~\citep{srivastava2013compete}.
In the future, we will try to adopt the trained CoPaNet to perform other tasks, such as object detection and segmentation.

\section*{Acknowledgement}
This work was supported in part by the Taiwan Ministry of Science and Technology (Grants MOST-106-2221-E-009-164-MY2 and MOST-105-2218-E-009-033). 

\bibliography{chang17}

\begin{thebibliography}{25}
\providecommand{\natexlab}[1]{#1}
\providecommand{\url}[1]{\texttt{#1}}
\expandafter\ifx\csname urlstyle\endcsname\relax
  \providecommand{\doi}[1]{doi: #1}\else
  \providecommand{\doi}{doi: \begingroup \urlstyle{rm}\Url}\fi

\bibitem[Abdi and Nahavandi(2016)]{abdi2016multi}
Masoud Abdi and Saeid Nahavandi.
\newblock Multi-residual networks.
\newblock \emph{arXiv preprint arXiv:1609.05672}, 2016.

\bibitem[Chang and Chen(2015)]{chang2015batch}
Jia-Ren Chang and Yong-Sheng Chen.
\newblock Batch-normalized maxout network in network.
\newblock \emph{arXiv preprint arXiv:1511.02583}, 2015.

\bibitem[Deng et~al.(2009)Deng, Dong, Socher, Li, Li, and
  Fei-Fei]{deng2009imagenet}
Jia Deng, Wei Dong, Richard Socher, Li-Jia Li, Kai Li, and Li~Fei-Fei.
\newblock Imagenet: A large-scale hierarchical image database.
\newblock In \emph{CVPR09}, pages 248--255. IEEE, 2009.

\bibitem[Goodfellow et~al.(2013)Goodfellow, Warde-Farley, Mirza, Courville, and
  Bengio]{goodfellow2013maxout}
Ian~J Goodfellow, David Warde-Farley, Mehdi Mirza, Aaron~C Courville, and
  Yoshua Bengio.
\newblock Maxout networks.
\newblock \emph{ICML (3)}, 28:\penalty0 1319--1327, 2013.

\bibitem[He et~al.(2015{\natexlab{a}})He, Zhang, Ren, and Sun]{he2015deep}
Kaiming He, Xiangyu Zhang, Shaoqing Ren, and Jian Sun.
\newblock Deep residual learning for image recognition.
\newblock \emph{arXiv preprint arXiv:1512.03385}, 2015{\natexlab{a}}.

\bibitem[He et~al.(2015{\natexlab{b}})He, Zhang, Ren, and Sun]{he2015delving}
Kaiming He, Xiangyu Zhang, Shaoqing Ren, and Jian Sun.
\newblock Delving deep into rectifiers: Surpassing human-level performance on
  imagenet classification.
\newblock In \emph{Proceedings of the IEEE International Conference on Computer
  Vision}, pages 1026--1034, 2015{\natexlab{b}}.

\bibitem[He et~al.(2016)He, Zhang, Ren, and Sun]{he2016identity}
Kaiming He, Xiangyu Zhang, Shaoqing Ren, and Jian Sun.
\newblock Identity mappings in deep residual networks.
\newblock \emph{arXiv preprint arXiv:1603.05027}, 2016.

\bibitem[Huang et~al.(2016{\natexlab{a}})Huang, Liu, and
  Weinberger]{huang2016densely}
Gao Huang, Zhuang Liu, and Kilian~Q Weinberger.
\newblock Densely connected convolutional networks.
\newblock \emph{arXiv preprint arXiv:1608.06993}, 2016{\natexlab{a}}.

\bibitem[Huang et~al.(2016{\natexlab{b}})Huang, Sun, Liu, Sedra, and
  Weinberger]{huang2016deep}
Gao Huang, Yu~Sun, Zhuang Liu, Daniel Sedra, and Kilian Weinberger.
\newblock Deep networks with stochastic depth.
\newblock \emph{arXiv preprint arXiv:1603.09382}, 2016{\natexlab{b}}.

\bibitem[Krizhevsky and Hinton(2009)]{krizhevsky2009learning}
Alex Krizhevsky and Geoffrey Hinton.
\newblock Learning multiple layers of features from tiny images, 2009.

\bibitem[Larsson et~al.(2016)Larsson, Maire, and
  Shakhnarovich]{larsson2016fractalnet}
Gustav Larsson, Michael Maire, and Gregory Shakhnarovich.
\newblock Fractalnet: Ultra-deep neural networks without residuals.
\newblock \emph{arXiv preprint arXiv:1605.07648}, 2016.

\bibitem[Lee et~al.(1999)Lee, Itti, Koch, and Braun]{lee1999attention}
Dale~K Lee, Laurent Itti, Christof Koch, and Jochen Braun.
\newblock Attention activates winner-take-all competition among visual filters.
\newblock \emph{Nature neuroscience}, 2\penalty0 (4):\penalty0 375--381, 1999.

\bibitem[Lin et~al.(2014)Lin, Chen, and Yan]{DBLP:journals/corr/LinCY13}
Min Lin, Qiang Chen, and Shuicheng Yan.
\newblock Network in network.
\newblock \emph{International Conference on Learning Representations},
  abs/1312.4400, 2014.
\newblock URL \url{http://arxiv.org/abs/1312.4400}.

\bibitem[Netzer et~al.(2011)Netzer, Wang, Coates, Bissacco, Wu, and
  Ng]{netzer2011reading}
Yuval Netzer, Tao Wang, Adam Coates, Alessandro Bissacco, Bo~Wu, and Andrew~Y
  Ng.
\newblock Reading digits in natural images with unsupervised feature learning.
\newblock In \emph{NIPS workshop on deep learning and unsupervised feature
  learning}, volume 2011, page~5. Granada, Spain, 2011.

\bibitem[Srivastava et~al.(2014{\natexlab{a}})Srivastava, Hinton, Krizhevsky,
  Sutskever, and Salakhutdinov]{srivastava2014dropout}
Nitish Srivastava, Geoffrey Hinton, Alex Krizhevsky, Ilya Sutskever, and Ruslan
  Salakhutdinov.
\newblock Dropout: A simple way to prevent neural networks from overfitting.
\newblock \emph{The Journal of Machine Learning Research}, 15\penalty0
  (1):\penalty0 1929--1958, 2014{\natexlab{a}}.

\bibitem[Srivastava et~al.(2013)Srivastava, Masci, Kazerounian, Gomez, and
  Schmidhuber]{srivastava2013compete}
Rupesh~K Srivastava, Jonathan Masci, Sohrob Kazerounian, Faustino Gomez, and
  J{\"u}rgen Schmidhuber.
\newblock Compete to compute.
\newblock In \emph{Advances in neural information processing systems}, pages
  2310--2318, 2013.

\bibitem[Srivastava et~al.(2015)Srivastava, Greff, and
  Schmidhuber]{srivastava2015training}
Rupesh~K Srivastava, Klaus Greff, and J{\"u}rgen Schmidhuber.
\newblock Training very deep networks.
\newblock In \emph{Advances in neural information processing systems}, pages
  2377--2385, 2015.

\bibitem[Srivastava et~al.(2014{\natexlab{b}})Srivastava, Masci, Gomez, and
  Schmidhuber]{srivastava2014understanding}
Rupesh~Kumar Srivastava, Jonathan Masci, Faustino Gomez, and J{\"u}rgen
  Schmidhuber.
\newblock Understanding locally competitive networks.
\newblock \emph{arXiv preprint arXiv:1410.1165}, 2014{\natexlab{b}}.

\bibitem[Szegedy et~al.(2015)Szegedy, Liu, Jia, Sermanet, Reed, Anguelov,
  Erhan, Vanhoucke, and Rabinovich]{szegedy2015going}
Christian Szegedy, Wei Liu, Yangqing Jia, Pierre Sermanet, Scott Reed, Dragomir
  Anguelov, Dumitru Erhan, Vincent Vanhoucke, and Andrew Rabinovich.
\newblock Going deeper with convolutions.
\newblock In \emph{Proceedings of the IEEE Conference on Computer Vision and
  Pattern Recognition}, pages 1--9, 2015.

\bibitem[Szegedy et~al.(2016)Szegedy, Ioffe, Vanhoucke, and
  Alemi]{szegedy2016inception}
Christian Szegedy, Sergey Ioffe, Vincent Vanhoucke, and Alex Alemi.
\newblock Inception-v4, inception-resnet and the impact of residual connections
  on learning.
\newblock \emph{arXiv preprint arXiv:1602.07261}, 2016.

\bibitem[Veit et~al.(2016)Veit, Wilber, and Belongie]{veit2016residual}
Andreas Veit, Michael~J Wilber, and Serge Belongie.
\newblock Residual networks behave like ensembles of relatively shallow
  networks.
\newblock In \emph{Advances in Neural Information Processing Systems}, pages
  550--558, 2016.

\bibitem[Wang et~al.(2015)Wang, Schaul, Hessel, van Hasselt, Lanctot, and
  de~Freitas]{wang2015dueling}
Ziyu Wang, Tom Schaul, Matteo Hessel, Hado van Hasselt, Marc Lanctot, and Nando
  de~Freitas.
\newblock Dueling network architectures for deep reinforcement learning.
\newblock \emph{arXiv preprint arXiv:1511.06581}, 2015.

\bibitem[Xie et~al.(2016)Xie, Girshick, Doll{\'a}r, Tu, and
  He]{xie2016aggregated}
Saining Xie, Ross Girshick, Piotr Doll{\'a}r, Zhuowen Tu, and Kaiming He.
\newblock Aggregated residual transformations for deep neural networks.
\newblock \emph{arXiv preprint arXiv:1611.05431}, 2016.

\bibitem[Zagoruyko and Komodakis(2016)]{zagoruyko2016wide}
Sergey Zagoruyko and Nikos Komodakis.
\newblock Wide residual networks.
\newblock \emph{arXiv preprint arXiv:1605.07146}, 2016.

\bibitem[Zhang et~al.(2016)Zhang, Li, Loy, and Lin]{zhang2016polynet}
Xingcheng Zhang, Zhizhong Li, Chen~Change Loy, and Dahua Lin.
\newblock Polynet: A pursuit of structural diversity in very deep networks.
\newblock \emph{arXiv preprint arXiv:1611.05725}, 2016.

\end{thebibliography}

\end{document}